\documentclass[10pt,twocolumn,letterpaper]{article}

\usepackage{cvpr}
\usepackage{times}
\usepackage{epsfig}
\usepackage{graphicx}
\usepackage{amsmath}
\usepackage{amssymb}
\usepackage{url}
\usepackage{authblk}

\newcommand{\myparagraph}[1]{\vspace{0.1em}\noindent\textbf{#1}}

\usepackage{tabularx}
\usepackage{booktabs}
\usepackage{rotating}
\usepackage[toc,page]{appendix}

\usepackage[pagebackref=true,breaklinks=true,letterpaper=true,colorlinks,bookmarks=false]{hyperref}

\newcommand{\timecnn}{$\tau_{\mbox{\tiny{CNN}}}$}
\newcommand{\timeinfer}{$\tau_{\mbox{\scriptsize{graph}}}$}
\newcommand{\videodata}{``MPII Video Pose''}
\newcommand{\deepcut}{\textit{DeepCut}}
\newcommand{\deepercut}{\textit{DeeperCut}}

\newcommand{\temporal}{\textit{temporal}}

\newcommand{\labeling}{\textit{label}}

\newcommand{\detection}{\textit{TD/BU}}
\newcommand{\regression}{\textit{BU}}

\newcommand{\detdistance}{\textit{det-distance}}
\newcommand{\deepmatch}{\textit{deepmatch}}
\newcommand{\siftdistance}{\textit{sift-distance}}

\newcommand{\bulong}{\textit{Bottom-Up}}
\newcommand{\tdbulong}{\textit{Top-Down/Bottom-Up}}
\newcommand{\tdlong}{\textit{Top-Down}}
\newcommand{\bushort}{\textit{BU}}
\newcommand{\tdbushort}{\textit{TD/BU}}
\newcommand{\bufull}{\textit{BU-full}}
\newcommand{\busparse}{\textit{BU-sparse}}
\newcommand{\tdshort}{\textit{TD}}
\newcommand{\spatprop}{\textit{SP}}

\newcommand{\bufullvid}{\textit{BU-full+temporal}}
\newcommand{\tdbuvid}{\textit{TD/BU+temporal}}

\newcommand{\busparsevid}{\textit{BU-sparse+temporal}}

\newcommand{\bvd}{\mathbf d}

\def\Sec{Sec\onedot}
\def\Tab{Tab\onedot}

\cvprfinalcopy 

\def\cvprPaperID{2880} 

\begin{document}

\title{ArtTrack: Articulated Multi-person Tracking in the Wild}

\author{Eldar Insafutdinov, Mykhaylo Andriluka, Leonid Pishchulin, Siyu Tang, \\\vspace{-0.2cm}Evgeny Levinkov, Bjoern Andres, Bernt Schiele\\
\vspace{0.5cm}
Max Planck Institute for Informatics\\
Saarland Informatics Campus\\
Saabr\"ucken, Germany\\
}

\maketitle

\begin{abstract}
  In this paper we propose an approach for articulated tracking of
  multiple people in unconstrained videos. Our starting point is a
  model that resembles existing architectures for single-frame pose
  estimation but is substantially faster. We achieve
  this in two ways: (1) by simplifying and sparsifying the body-part
  relationship graph and leveraging recent methods for faster
  inference, and (2) by offloading a substantial share of computation
  onto a feed-forward convolutional architecture that is able to
  detect and associate body joints of the same person even in
  clutter. We use this model to generate proposals for body joint
  locations and formulate articulated tracking as spatio-temporal
  grouping of such proposals. This allows to jointly solve the
  association problem for all people in the scene by propagating evidence
  from strong detections through time and enforcing constraints that
  each proposal can be assigned to one person only. We report results
  on a public ``MPII Human Pose'' benchmark and on a new \videodata~dataset of image sequences
  with multiple people. We demonstrate that our model achieves
  state-of-the-art results while using only a fraction of time and is
  able to leverage temporal information to improve state-of-the-art
  for crowded scenes\footnote{The models and the \videodata~dataset are available at \url{pose.mpi-inf.mpg.de/art-track}.}.
\end{abstract}

\section{Introduction}

This paper addresses the task of articulated human pose tracking in monocular video. We focus on 
scenes of realistic complexity that often include fast motions, large variability in
appearance and clothing, and person-person occlusions. A successful approach must thus identify the number of people in each video frame, determine locations of the joints
of each person and associate the joints over time.

\begin{figure}
  \centering
  \begin{tabular}{c c}

  \includegraphics[width=0.48\linewidth]{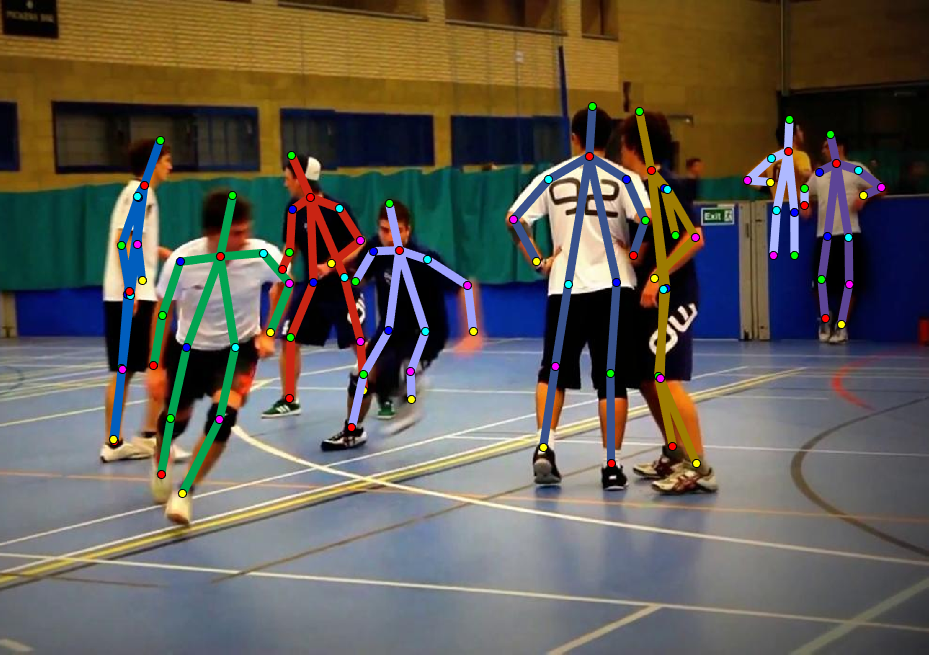}&
  \includegraphics[width=0.48\linewidth]{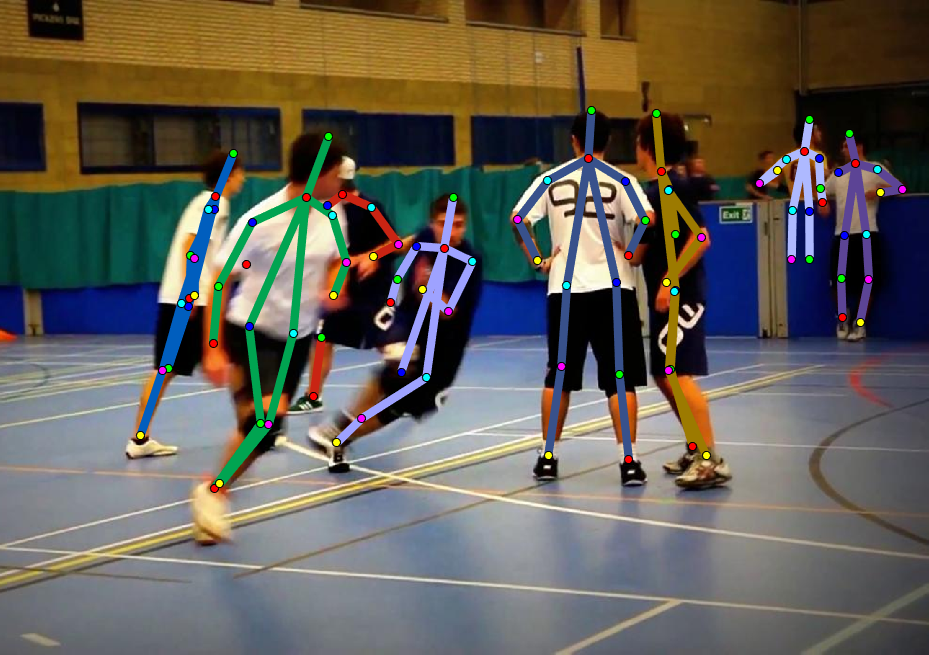}\\

  \end{tabular} 
  \caption{Example articulated tracking results of our approach.}
 \vspace{-1.0em}
  \label{fig:teaser}
\end{figure}

One of the key challenges in such scenes is that people might overlap and only a subset of joints of
the person might be visible in each frame either due to person-person occlusion or truncation by
image boundaries (\cf.~Fig.~\ref{fig:teaser}). Arguably, resolving such cases correctly requires
reasoning beyond purely geometric information on the arrangement of body joints in the image, and
requires incorporation of a variety of image cues and joint modeling of several persons. 

\tabcolsep 1.5pt

The design of our model is motivated by two factors. We would like to leverage bottom-up end-to-end
learning to directly capture image information. At the same time we aim to address a complex
multi-person articulated tracking problem that does not naturally lend itself to an
end-to-end prediction task and for which training data is not available in the amounts usually
required for end-to-end learning.

To leverage the available image information we learn a model for associating a body joint to a
specific person in an end-to-end fashion relying on a convolutional network. We then incorporate
these part-to-person association responses into a framework for jointly reasoning about assignment
of body joints within the image and over time. To that end we use the graph partitioning formulation
that has been used for people tracking and pose estimation in the past
\cite{tang-2015,pishchulin16cvpr}, but has not been shown to enable articulated people tracking.

To facilitate efficient inference in video we resort to fast inference
methods based on local combinatorial optimization
\cite{levinkov16arxiv} and aim for a sparse model that keeps the
number of connections between variables to a minimum. As we
demonstrate, in combination with feed-forward reasoning for
joint-to-person association this allows us to achieve substantial speed-ups
compared to state-of-the-art \cite{insafutdinov16eccv} while maintaining the same level of
accuracy.



The main contribution of this work is a new articulated tracking model that operates by bottom-up assembly of part detections within
each frame and over time. In contrast to \cite{gkioxari16eccv,Pfister15} this model is
suitable for scenes with an unknown number of subjects and 
reasons jointly across multiple people incorporating inter-person
exclusion constraints and propagating strong observations to
neighboring frames.
%

Our second contribution is a formulation for single-frame pose estimation that relies on a sparse
graph between body parts and a mechanism for generating body-part proposals conditioned on a person's
location. This is in contrast to state-of-the-art approaches \cite{pishchulin16cvpr,insafutdinov16eccv} that
perform expensive inference in a full graph and rely on generic bottom-up proposals. We
demonstrate that a sparse model with a few spatial edges performs competitively with a
fully-connected model while being much more efficient. Notably, a simple model that operates in
top-down/bottom-up fashion exceeds the performance of a fully-connected model while being $24$x
faster at inference time (cf.~\Tab.~\ref{tab:mpii-single-frame:sota}). This is due to offloading of a large
share of the reasoning about body-part association onto a feed-forward convolutional architecture.

Finally, we contribute a new challenging dataset for evaluation of
articulated body joint tracking in crowded realistic environments with
multiple overlapping people. 



\myparagraph{Related work.} Convolutional networks have emerged as an effective approach to
localizing body joints of people in
images \cite{tompson14nips,wei16cvpr,newell16eccv,insafutdinov16eccv}
and have also been extended for joint estimation of body
configurations over time \cite{gkioxari16eccv}, and 3D pose estimation 
in outdoor environments in multi-camera setting \cite{elhayek15cvpr,elhayek16pami}.

Current approaches are increasingly effective for estimating body
configurations of single
people~\cite{tompson14nips,wei16cvpr,newell16eccv,bulat16eccv,gkioxari16eccv}
achieving high accuracies on this task, but are still failing on fast
moving and articulated limbs. More complex recent models jointly
reason about entire scenes~\cite{pishchulin16cvpr,insafutdinov16eccv,Iqbal_ECCVw2016},
but are too complex and inefficient to directly generalize to image sequences. 
Recent feed-forward models are able
to jointly infer body joints of the same person and even operate over
time~\cite{gkioxari16eccv} but consider isolated persons only and do
not generalize to the case of multiple overlapping
people. Similarly, \cite{Charles16,Pfister15} consider a simplified
task of tracking upper body poses of isolated upright individuals.

%
%
We build on recent CNN detectors \cite{insafutdinov16eccv} that are
effective in localizing body joints in cluttered scenes
and explore different mechanisms for assembling the joints into
multiple person configurations. To that end we rely on a graph
partitioning approach closely related
to~\cite{tang-2015,pishchulin16cvpr,insafutdinov16eccv}. In contrast
to~\cite{tang-2015} who focus on pedestrian tracking,
and~\cite{pishchulin16cvpr,insafutdinov16eccv} who perform single
frame multi-person pose estimation, we solve a more complex
problem of articulated multi-person pose tracking. 

Earlier approaches to articulated pose tracking in monocular videos rely on hand-crafted image
representations and focus on simplified tasks, such as tracking upper body poses of
frontal isolated people~\cite{sapp11cvpr,weiss13nips,Tokola_ICCV2013,Cherian14}, or tracking walking
pedestrians with little degree of articulation~\cite{Andriluka:CVPR08,andriluka10}. In contrast, we
address a harder problem of multi-person articulated pose tracking and do not make assumptions about
the type of body motions or activities of people.
Our approach is closely
related to \cite{iqbal17cvpr} who propose a similar formulation based on graph partitioning. Our
approach differs from \cite{iqbal17cvpr} primarily in the type of body-part proposals and the
structure of the spatio-temporal graph. In our approach we introduce a person-conditioned model that
is trained to associate body parts of a specific person already at the detection stage. This is in
contrast to the approach of \cite{iqbal17cvpr} that relies on the generic body-part detectors \cite{insafutdinov16eccv}.

%
% ...

\myparagraph{Overview. } Our model consists of the two components: (1) a convolutional network for
generating body part proposals and (2) an approach to group the proposals into spatio-temporal
clusters. In Sec.~\ref{section:tracking-general} we introduce a general formulation
for multi-target tracking that follows \cite{tang-2015} and allows us to define pose estimation and
articulated tracking in a unified framework. We then describe the details of our articulated
tracking approach in Sec.~\ref{section:tracking-pose}, and introduce two variants of our
formulation: bottom-up (\bushort) and top-down/bottom-up (\tdbushort). We present experimental
results in Sec.~\ref{seq:experiments}.

\section{Tracking by Spatio-temporal Grouping}
\label{section:tracking-general}

Our body part detector generates a set of proposals $D = \{\bvd_i\}$
for each frame of the video. Each proposal is given by $\bvd_i =
(t_i, d^{pos}_{i}, \pi_i, \tau_i)$, where $t_i$ denotes the index
of the video frame, $d^{pos}_{i}$ is the spatial
location of the proposal in image coordinates, $\pi_i$ is the probability of correct
detection, and $\tau_i$ is the type of the body joint (\eg ankle or shoulder). 

Let $G = (D,E)$ be a graph whose nodes $D$ are the joint detections in a
video and whose edges $E$ connect pairs of detections that hypothetically correspond to the 
same target.

The output of the tracking algorithm is a subgraph $G'=(D', E')$ of $G$, where $D'$ is a subset of nodes
after filtering redundant and erroneous detections and $E'$ are edges linking nodes
corresponding to the same target. We specify $G'$ via binary variables $x \in \{0,1\}^D$
and $y \in \{0,1\}^E$ that define subsets of edges and nodes included in $G'$. In particular each
track will correspond to a connected component in $G'$.

As a general way to introduce constraints on edge configurations that correspond to a valid tracking
solution we introduce a set $Z \subseteq \{0,1\}^{D \cup E}$ and define a combination of edge and
node indicator variables to be feasible if and only if $(x, y) \in Z$. An example of a constraint
encoded through $Z$ is that endpoint nodes of an edge included by $y$ must also be included by
$x$. Note that the variables $x$ and $y$ are coupled though $Z$. Moreover, assuming that $(x, y) \in Z$ we are
free to set components of $x$ and $y$ independently to maximize the tracking objective.

Given image observations we compute a set of features for each node and edge in the graph. We denote
such node and edge features as $f$ and $g$ respectively. Assuming independence of the feature
vectors the conditional probability of indicator functions $x$ of nodes and $y$ of edges given
features $f$ and $g$ and given a feasible set $Z$ is given by
\begin{align}
\label{eq:joint-probability}
\hspace{-0.7em} p(x, y | f, g, Z) \propto p(Z | x, y) \prod_{d \in D} p(x_d | f^d) \prod_{e \in E} p(y_e | g^e),
\end{align}
\noindent where $p(Z | x, y)$ assigns a constant non-zero probability to every feasible solution and
is equal to zero otherwise. Minimizing the negative log-likelihood of Eq.~\ref{eq:joint-probability}
is equivalent to solving the following integer-linear program: 
\begin{align}
\label{eq:ilp}
\min_{\begin{array}{c}\scriptstyle (x, y) \in Z\end{array}}\sum_{d \in D} c_d x_d + \sum_{e \in E} d_e y_e
\enspace ,
\end{align}
\noindent where $c_d=\log\frac{p(x_d=1|f^d)}{p(x_d=0|f^d)}$ is the cost of retaining $d$ as part of
the solution, and $d_e=\log\frac{p(y_e=1|g^e)}{p(y_e=0|g^e)}$ is the cost of assigning the
detections linked by an edge $e$ to the same track.

We define the set of constraints $Z$ as in \cite{tang-2015}:
\begin{align}
  & \quad \forall e=vw \in E:
  \quad y_{vw} \leq x_v \label{eq:mc-node-edge-1}\\
  & \quad \forall e=vw \in E:
  \quad y_{vw} \leq x_w \label{eq:mc-node-edge-2}\\
  & \quad \forall C \in \mathrm{cycles}(G)\ \forall e \in C:\ \nonumber \\
  & \qquad \quad (1 - y_e) \leq \hspace{-2ex} \sum_{e' \in C \setminus \{e\}} \hspace{-2ex} (1 - y_{e'}) \label{eq:mc-node-edge-3}
\end{align} 

Jointly with the objective in Eq.~\ref{eq:ilp} the constraints
(\ref{eq:mc-node-edge-1})-(\ref{eq:mc-node-edge-3}) define an instance
of the minimum cost subgraph multicut problem \cite{tang-2015}. The
constraints (\ref{eq:mc-node-edge-1}) and (\ref{eq:mc-node-edge-2})
ensure that assignment of node and edge variables is consistent. The
constraint (\ref{eq:mc-node-edge-3}) ensures that for every two nodes
either all or none of the paths between these nodes in graph $G$ are
contained in one of the connected components of subgraph $G'$. This
constraint is necessary to unambigously assign person identity
to a body part proposal based on its membership in a specific connnected
component of $G'$. 


\section{Articulated Multi-person Tracking}
\label{section:tracking-pose}

\begin{figure}
  \centering
  \setlength\tabcolsep{8pt}
  \begin{tabular}{c c c}
  \includegraphics[height=0.20\linewidth]{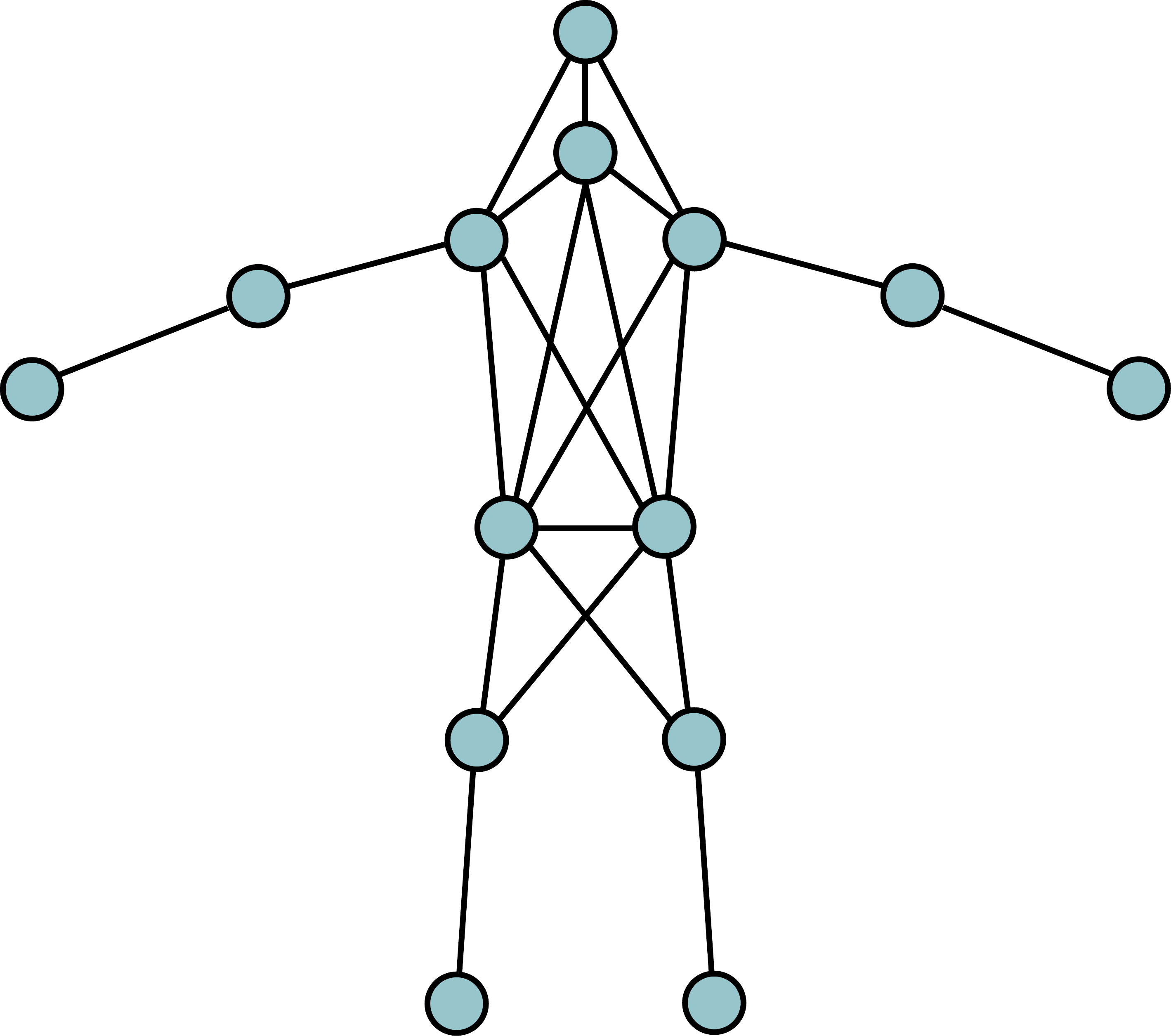}&
  \includegraphics[height=0.25\linewidth]{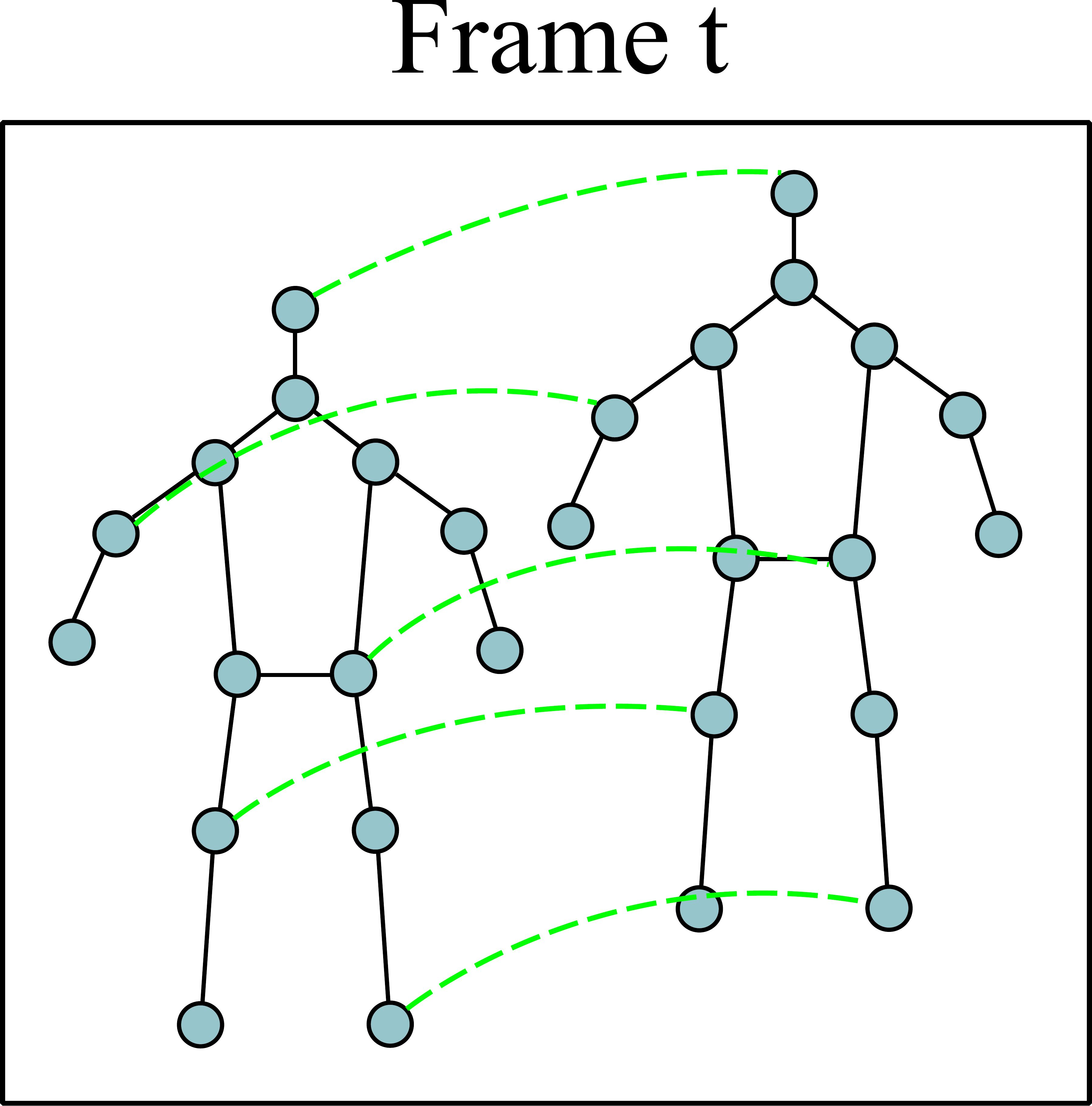}&
  \includegraphics[height=0.35\linewidth]{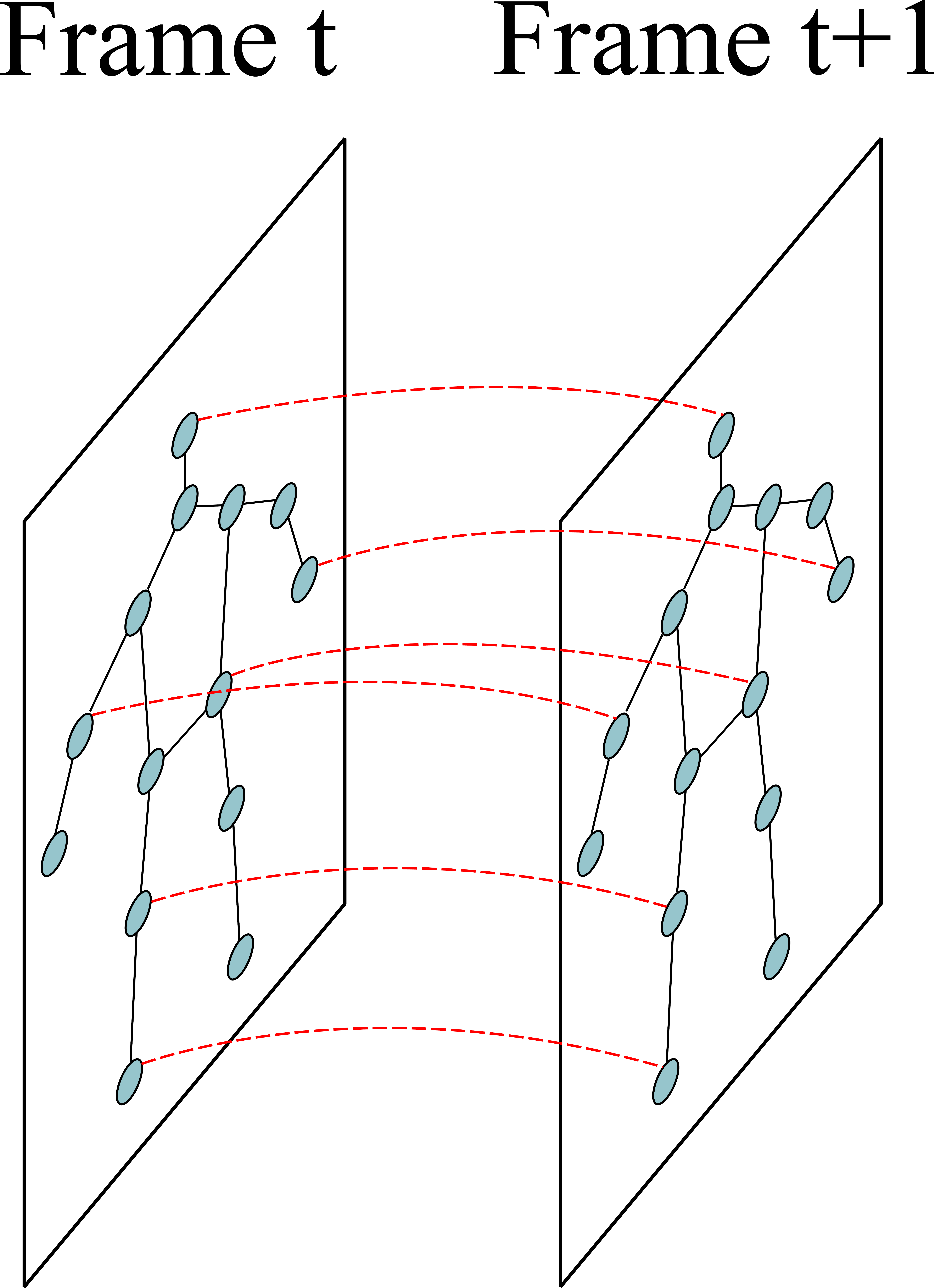}\\
  (a) & (b) & (c)\\
    \end{tabular}
    \caption{Visualization of (a) sparse connectivity, (b)
      attractive-repulsive edges and (c) temporal edges in our
      model. We show only a subset of attractive/repulsive and
      temporal edges for clarity.}
  \label{fig:edge_types}
\end{figure}

\begin{figure*}
  \centering
  \begin{tabular}{c c c c c c}
  \includegraphics[height=0.35\linewidth]{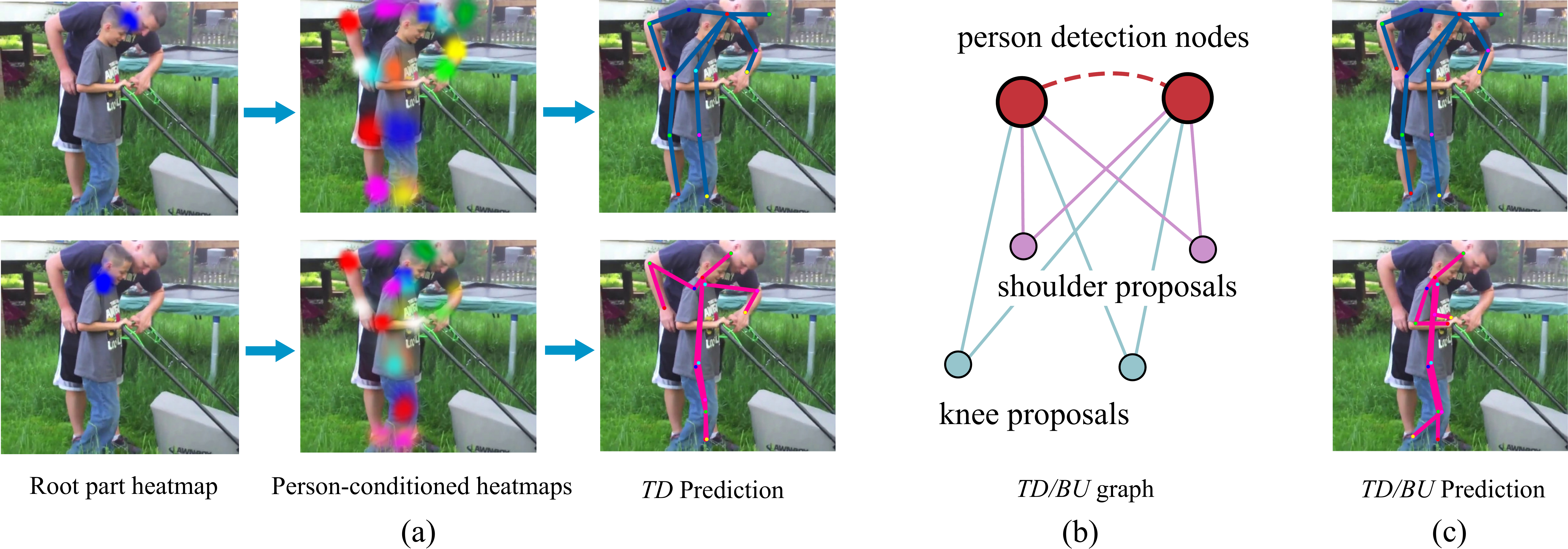}&
  \end{tabular}
\caption{\textbf{(a)} Processing stages of the \tdlong{} model shown for an example with
  significantly overlapping people. Left:  Heatmaps for the chin (=root part) used to condition the CNN on the location of the person in the back (top) and in the
front (bottom). Middle: Output heatmaps for all body parts, notice the ambiguity in estimates of the arms of the front person. Right: \tdshort{} predictions for each person. \textbf{(b)} Example of the \tdbulong~graph. Red dotted line represents the must-cut constraint. Note that body part proposals of different type are connected to person nodes
   	but not between each other.	\textbf{(c)} \tdbulong{} predictions. Notice that the \tdbushort{} inference correctly assigns the forearm joints of the frontal person.}
   \vspace{-1.0em}
  \label{fig:top_down_bottom_up_heatmaps}
\end{figure*}

In Sec.~\ref{section:tracking-general} we introduced a general framework for multi-object tracking
by solving an instance of the subgraph multicut problem. The subgraph multicut problem is NP-hard, but 
recent work \cite{tang-2015,levinkov16arxiv} has shown that efficient approximate inference is possible with local search methods.
The framework allows for a variety of graphs and connectivity patterns. Simpler
connectivity allows for faster and more efficient processing at the cost of ignoring 
some of the potentially informative dependencies between model variables. Our goal is to design a model that is efficient, with as
few edges as possible, yet effective in crowded scenes, and that allows us to model temporal
continuity and inter-person exclusion. 
Our articulated tracking approach proceeds by constructing a graph $G$ that couples
body part proposals within the same frame and across neighboring frames. In general the graph $G$
will have three types of edges: (1) \textit{cross-type} edges shown in
  Fig.~\ref{fig:edge_types} (a) and Fig.~\ref{fig:top_down_bottom_up_heatmaps} (b) that connect two parts of different types, (2) \textit{same-type} edges shown in Fig.~\ref{fig:edge_types} (b) that connect two nodes of the same type in the same image, and (3) \textit{temporal} edges shown in Fig.~\ref{fig:edge_types} (c) that connect nodes in the neighboring frames.

We now define two variants of our model that we denote as $\bulong$ (\bushort) and
$\tdbulong$ (\tdbushort). In the $\bushort$ model the body part proposals are generated with our publicly
available convolutional part
detector~\cite{insafutdinov16eccv}\footnote{http://pose.mpi-inf.mpg.de/}. In the $\tdbushort$ model we
substitute these generic part detectors with a new convolutional body-part detector that is trained
to output consistent body configurations conditioned on the person location. This alows to further
reduce the complexity of the model graph since the task of associating body parts is addressed
within the proposal mechanism. As we show in Sec.~\ref{seq:experiments} this leads to considerable
gains in performance and allows for faster inference. Note that the \bushort~and \tdbushort~models
have identical \textit{same-type} and \textit{temporal} pairwise terms, but differ in the form of 
\textit{cross-type} pairwise terms, and the connectivity of the nodes in $G$. For both models we 
rely on the solver from \cite{levinkov16arxiv} for inference.

\subsection{Bottom-Up Model (\bushort).} 
\label{sec:bu}
For each body part proposal $\bvd_i$ the detector outputs image location,
probability of detection $\pi_i$, and a label $\tau_i$ that indicates
the type of the detected part (\eg shoulder or ankle). We
directly use the probability of detection to derive the unary costs in
Eq.~\ref{eq:ilp} as $c_{d_i}=\log(\pi_i/(1-\pi_i))$. Image features
$f^d$ in this case correspond to the image representation generated by
the convolutional network.

We consider two connectivity patterns for nodes in the graph $G$. We
either define edges for every pair of proposals which results in a
fully connected graph in each image. Alternatively we obtain a sparse
version of the model by defining edges for a subset of part types only
as is shown in Fig.~\ref{fig:edge_types} (a). The rationale behind the
sparse version is to obtain a simpler and faster version of the model
by omitting edges between parts that carry little information about
each other's image location (\eg left ankle and right arm).

\myparagraph{Edge costs.} In our $\bulong$ model the cost of the edges
$d_e$ connecting two body part detections $\bvd_i$ and $\bvd_j$ is
defined as a function of the detection types $\tau_i$ and $\tau_j$.
Following~\cite{insafutdinov16eccv} we thus train for each pair of
part types a regression function that predicts relative image location of
the parts in the pair. The cost $d_e$ is given by the output of the logistic
regression given the features computed from offset and angle of the
predicted and actual location of the other joint in the pair. We refer
to \cite{insafutdinov16eccv} for more details on these pairwise terms.

Note that our model generalizes \cite{tang-2015} in that the edge cost
depends on the type of nodes linked by the edge. It also
generalizes \cite{pishchulin16cvpr,insafutdinov16eccv} by allowing $G$
to be sparse. This is achieved by reformulating the model with a more
general type of cycle constraint (\ref{eq:mc-node-edge-3}), in
contrast to simple triangle inequalities used in
\cite{pishchulin16cvpr,insafutdinov16eccv}\footnote{See Sec.~2.1 in
  \cite{pishchulin16cvpr}}.





\subsection{Top-Down/Bottom-up Model (\tdbushort)}
\label{sec:person-cond}

\begin{figure*}
  \centering
  \begin{tabular}{c c c c c c}
  \includegraphics[height=0.200\linewidth]{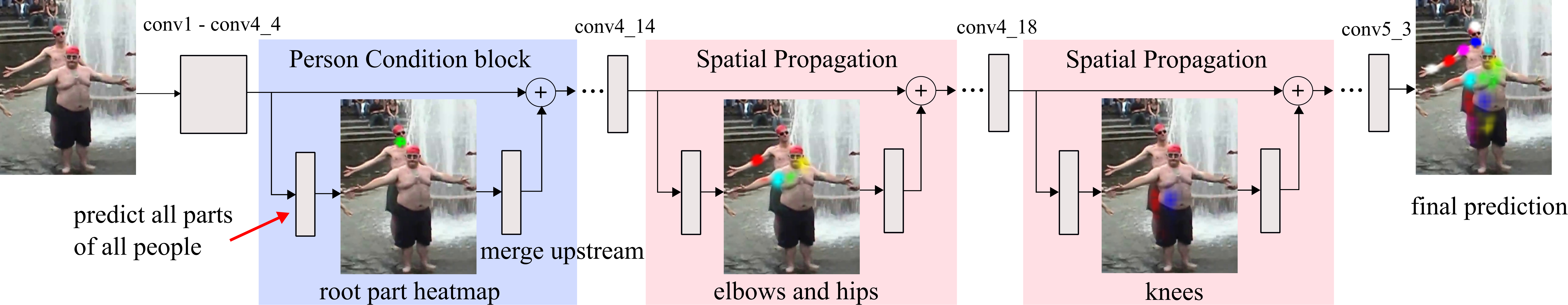}&
  \end{tabular}
   \caption{CNN architecture based on ResNet-101 for computing person conditioned proposals and pairwise terms. \spatprop{} block for shoulders at \textit{conv\_4\_8} is omitted for clarity.} 
   \vspace{-1.0em}
  \label{fig:person_condition_arch}
\end{figure*}

We now introduce a version of our model that operates by first generating body part proposals
conditioned on the locations of people in the image and then performing joint reasoning to group
these proposals into spatio-temporal clusters corresponding to different people. We follow the
intuition that it is considerably easier to identify and detect individual people (e.g.~by
detecting their heads) compared to correctly associating body parts such as ankles and wrists to
each person. We select person's head as a root part that is responsible for representing the person
location, and delegate the task of identifying body parts of the person corresponding to a head
location to a convolutional network.

The structure of \tdbushort{} model is illustrated in
Fig.~\ref{fig:top_down_bottom_up_heatmaps} (b) for the simplified case
of two distinct head detections. Let us denote the set of all root
part detections as $D^{root} = \{d_i^{root}\}$. For each pair of the
root nodes we explicitly set the corresponding edge indicator
variables $y_{d_j^{root},d_k^{root}}=0$. This implements a
``must-not-link'' constraint between these nodes, and in combination
with the cycle inequality (\ref{eq:mc-node-edge-3}) implies that each
proposal can be connected to one of the ``person nodes'' only. The
cost for an edge connecting detection proposal $\bvd_k$ and a ``person
node'' $d_i^{root}$ is based on the conditional distribution
$p_{d_k^c}(d_k^{pos}| d_i^{root})$ generated by the convolutional
network. The output of such network is a set of conditional
distributions, one for each node type. We augment the graph $G$ with
attractive/repulsive and temporal terms as described in
Sec.~\ref{subsection:attractive} and Sec.~\ref{subsection:temporal}
and set the unary costs for all indicator variables $x_d$ to a
constant. Any proposal not connected to any of the root nodes is
excluded from the final solution. We use the solver from
\cite{levinkov16arxiv} for consistency, but a simpler KL-based solver
as in \cite{tang-2015,keuper-2015a} could be used as well since the
\tdbushort~model effectively ignores the unary variables
$x_d$. The processing stages of \tdbushort{} model are shown in
Fig.~\ref{fig:top_down_bottom_up_heatmaps}. Note that the body-part
heatmaps change depending on the person-identity signal
provided by the person's neck, and that the bottom-up step was able to
correct the predictions on the forearms of the front person.

\myparagraph{Implementation details.} 
For head detection, we use a version of our model that
contains the two head parts (neck and head top). This makes our
\tdbushort{} model related to the hierarchical model defined in
\cite{insafutdinov16eccv} that also uses easier-to-detect parts to
guide the rest of the inference process. However here we replace
all the stages in the hierarchical inference except the first one
with a convolutional network.
%

The structure of the convolutional network used to generate person-conditioned proposals is shown on
Fig.~\ref{fig:person_condition_arch}. The network uses the ResNet-101 from
\cite{he2015deep} that we modify to bring the stride of the network down to
8 pixels~\cite{insafutdinov16eccv}.
%
%
The network generates predictions for all body parts after the \textit{conv4\_4} block. We use the
cross-entropy binary classification loss at this stage to predict the part heatmaps. 
At each training iteration we forward pass an image with multiple people potentially in close
proximity to each other. We select a single person from the image and condition the network on the
person's neck location by zeroing out the heatmap of the neck joint outside the ground-truth region. 
We then pass the neck heatmap through a  convolutional layer to match the dimensionality of the feature channels
and add them to the main stream of the ResNet.  We finally add a joint prediction layer at the end
of the network with a loss that considers predictions to be correct only if they correspond to the body
joints of the selected person.
%
%

\myparagraph{Spatial propagation (\spatprop).} In our network the person identity signal is provided by the
location of the head. In principle the receptive field size of the network is large enough to propagate this signal to all
body parts.
However we found that it is useful to introduce an additional mechanism to propagate the person
identity signal. To that end we inject intermediate
supervision layers for individual body parts arranged in the order of kinematic proximity to the
root joint (Fig.~\ref{fig:person_condition_arch}). We place prediction layers for shoulders at
\textit{conv4\_8}, for elbows and hips at \textit{conv4\_14} and for knees at
\textit{conv4\_18}.
We empirically found that such an explicit
form of spatial propagation significantly improves performance on joints such as ankles,
that are typically far from the head in the image space (see Tab.~\ref{tab:mpii-single-frame-val:detection} for details).

\myparagraph{Training.} We use Caffe's~\cite{jia2014caffe} ResNet implementation and
initialize from the ImageNet-pre-trained models. Networks are trained on the MPII Human Pose dataset~\cite{andriluka14cvpr} with SGD
for 1M iterations with stepwise learning rate (lr=0.002 for 400k, lr=0.0002 for 300k and lr=0.0001 for 300k).

\subsection{Attractive/Repulsive Edges}
\label{subsection:attractive}

Attractive/repulsive edges are defined between two proposals of the same type within the same
image. The costs of these edges is inversely-proportional to distance
\cite{insafutdinov16eccv}. 
The decision to group two nodes is made based on the evidence from the
entire image, which is in contrast to typical non-maximum suppression
based on the state of just two detections. Inversely, these edges
prevent grouping of multiple distant hypothesis of the same type, \eg
prevent merging two heads of different people.


\subsection{Temporal Model}
\label{subsection:temporal}

Regardless of the type of within frame model (\bushort{} or \tdbushort) we rely on the same type of
temporal edges that connect nodes of the same type in adjacent frames. We derive the costs for such
temporal edges via logistic regression. Given the feature vector $g_{ij}$ the probability that the
two proposals $\bvd_i$ and $\bvd_j$ in adjacent frames correspond to the same body part is given by:
%
%
$ p(y_{ij} = 1|g_{ij}) = 1/(1 + \exp(-\langle \omega_{t}, g_{ij} \rangle))$,
where $g_{ij} = (\Delta_{ij}^{L2}, \Delta_{ij}^{Sift}, \Delta_{ij}^{DM}, \tilde\Delta_{ij}^{DM})$, and 
 $\Delta_{ij}^{L2}= \|d^{pos}_{i}$ - $d^{pos}_{j}\|_2$, $\Delta_{ij}^{Sift}$ is Euclidean distance
between the SIFT descriptors computed at $d^{pos}_{i}$ and $d^{pos}_{j}$, and $\Delta_{ij}^{DM}$ and
$\tilde\Delta_{ij}^{DM}$ measure the agreement with the dense motion field computed with the
DeepMatching approach of \cite{weinzaepfelhal00873592}. 

For SIFT features we specify the location of the detection proposal, but rely on SIFT to identify
the local orientation. In cases with multiple local maxima in orientation estimation we compute SIFT
descriptor for each orientation and set $\Delta_{ij}^{Sift}$ to the minimal distance among all pairs
of descriptors. We found that this makes the SIFT distance more robust in the presence of rotations
of the body limbs.

We define the features $\Delta_{ij}^{DM}$ and $\tilde\Delta_{ij}^{DM}$ as in \cite{tangbmtt}.
Let $R_i = R(\bvd_i)$ be an squared image region centered on the part proposal $\bvd_i$. We define
$\Delta_{ij}^{DM}$ as a ratio of the number of point correspondences between the regions $R_i$ and
$R_j$ and the total number of point correspondences in either of them. Specifically, let $C =
\{c^k|k=1, \ldots, K\}$ be a set of point correspondences between the two images computed with
DeepMatching, where $c^k = (c^k_1, c^k_2)$ and $c^k_1$ and $c^k_2$ denote the corresponding points
in the first and second image respectively. Using this notation we define:
\begin{align}
\Delta_{ij}^{DM} = \frac{|\{c_k| c^k_1 \in R_i \land c^k_2 \in R_j\}|}{|\{c_k| c^k_1 \in R_i\}| + |\{c_k| c^k_2 \in R_j\}|}.
\end{align}
The rationale behind computing $\Delta_{ij}^{DM}$ by aggregating across multiple correspondences is
to make the feature robust to outliers and to inaccuracies in body part
detection. $\tilde\Delta_{ij}^{DM}$ is defined analogously, but using the DeepMatching
correspondences obtained by inverting the order of images.

\myparagraph{Discussion.} As we demonstrate in Sec.~\ref{seq:experiments}, we found the set of
features described above to be complementary to each other. Euclidean distance between proposals is
informative for finding correspondences for slow motions, but fails for faster motions and in the
presence of multiple people. DeepMatching is usually effective in finding corresponding regions
between the two images, but occasionally fails in the case of sudden background changes due to fast
motion or large changes in body limb orientation. In these cases SIFT is often still able to provide
a meaningful measure of similarity due to its rotation invariance.


% ... 






\section{Experiments}
\label{seq:experiments}


\subsection{Datasets and evaluation measure}
\myparagraph{Single frame.} We evaluate our single frame models on
the MPII Multi-Person dataset~\cite{andriluka14cvpr}. 
We report all intermediate results on a validation set of $200$ images
sampled uniformly at random (MPII Multi-Person Val), while major
results and comparison to the state of the art are reported on the
test set.

\myparagraph{Video.} In order to evaluate video-based models we
introduce a novel \videodata~dataset\footnote{Dataset is available at \url{pose.mpi-inf.mpg.de/art-track}.}.
 To this end we manually selected challenging
keyframes from MPII Multi-Person dataset. Selected
keyframes represent crowded scenes with highly articulated
people engaging in various dynamic activities. In addition to each
keyframe, we include +/-$10$ neighboring frames from the
corresponding publicly available video sequences, and annotate every
second frame\footnote{The annotations in the original key-frame are kept unchanged.}.
%
Each body pose was annotated following the standard annotation
procedure~\cite{andriluka14cvpr}, while maintaining person identity
throughout the sequence. In contrast to MPII Multi-Person where some
frames may contain non-annotated people, we annotate all people
participating in the activity captured in the video, and add ignore
regions for areas that contain dense crowds (e.g. static spectators in
the dancing sequences).  In total, our dataset consists of $28$
sequences with over $2,000$ annotated poses.


\myparagraph{Evaluation details.} 
The average precision (AP) measure~\cite{pishchulin16cvpr} is used for evaluation of pose estimation
accuracy. For each algorithm we also report run time \timecnn~of the proposal generation
and \timeinfer~of the graph partitioning stages. All time measurements were conducted on a single
core Intel Xeon $2.70$GHz. Finally we also evaluate tracking perfomance using standard MOTA metric \cite{Bernardin:2008:CLE}.

%
%
Evaluation on our \videodata~dataset is performed on the full
frames using the publicly available evaluation kit
of~\cite{andriluka14cvpr}. On MPII Multi-Person we follow the official
evaluation
protocol\footnote{http://human-pose.mpi-inf.mpg.de/\#evaluation} and
evaluate on groups using the provided rough group location and scale.

\subsection{Single-frame models}
\tabcolsep 1.5pt
\begin{table}[tbp]
 \scriptsize
  \centering
  \begin{tabular}{@{} l c ccc ccc c | c  cc@{}}
    \toprule
    Setting& Head   & Sho  & Elb & Wri & Hip & Knee & Ank & AP & \timecnn & \timeinfer\\
    \midrule

    




     \bufull, \labeling & 90.0  & 84.9  & 71.1  & 58.4  & 69.7  & 64.7 & 54.7 & 70.5 &   \textbf{0.18} & 3.06 \\ 
     \bufull            & 91.2  & 86.0  & \textbf{72.9}  & 61.5  & 70.4  & 65.4 & 55.5 & 71.9 &   \textbf{0.18} & 0.38 \\ 
     \busparse          & 91.1  & \textbf{86.5}  & 70.7  & 58.1  & 69.7  & 64.7 & 53.8 & 70.6 &   \textbf{0.18} & 0.22 \\ 
     \midrule
     \tdbushort{} + \spatprop & \textbf{92.2}  & 86.1  & 72.8  & \textbf{63.0}  & \textbf{74.0}  &  \textbf{66.2} & \textbf{58.4} & \textbf{73.3} &  0.94{\color{red}\footnotemark[7]} & \textbf{0.08} \\ 
     

    \bottomrule
  \end{tabular}
 \vspace{0.75em}
\caption[]{Effects of various variants of $\regression$ model on pose estimation performance (AP) on
  MPII Multi-Person Val and comparison to the best variant of \tdbushort model. 
}
\label{tab:mpii-single-frame-val:regression}
\end{table}

We compare the performance of different variants of our \bulong~(\bushort) and
\tdbulong~(\tdbushort) models introduced in Sec.~\ref{sec:bu} and Sec.~\ref{sec:person-cond}.  For
\bushort~we consider a model that (1) uses a fully-connected graph with up to $1,000$ detection
proposals and jointly performs partitioning \textit{and} body-part labeling similar
to~\cite{insafutdinov16eccv} (\bufull, \labeling); (2) is same as (1), but labeling of detection
proposals is done based on detection score (\bufull); (3) is same as (2), but uses a sparsely-connected
graph (\busparse). The results are shown in
Tab.~\ref{tab:mpii-single-frame-val:regression}\footnote{\label{cnntime}Our current implementation
  of \tdbushort~operates on the whole image when computing person-conditioned proposals and computes
  the proposals sequentially for each person. More efficient implementation would only compute the
  proposals for a region surrounding the person and run multiple people in a single batch. Clearly
  in cases when two people are close in the image this would still process the same image region
  multiple times. However the image regions far from any person would be excluded from processing
  entirely. On average we expect similar image area to be processed during proposal generation stage
  in both \tdbushort~and \busparse,~and expect the runtimes \timecnn~to be comparable for both
  models.}.  \bufull, \labeling{} achieves $70.5$\% AP with a median inference run-time
\timeinfer~of $3.06$ s/f. \bufull{} achieves $8\times$ run-time reduction ($0.38$ vs. $3.06$ s/f):
pre-labeling detection candidates based on detection score significantly reduces the number of
variables in the problem graph. Interestingly, pre-labeling also improves the performance ($71.9$
vs. $70.5$\% AP): some of the low-scoring detections may complicate the search for an optimal
labeling. $\busparse$ further reduces run-time ($0.22$ vs. $0.38$ s/f), as
it reduces the complexity of the initial problem by sparsifying the graph, at a price of a drop in
performance ($70.6$ vs. $71.9$\% AP).

In Tab.~\ref{tab:mpii-single-frame-val:detection} we compare the variants of the \tdbushort~model.
Our $\tdshort$~approach achieves $71.7$\% AP, performing on par with a more complex $\bufull$. 
Explicit spatial propagation
(\tdshort{}+\spatprop ) further improves the results ($72.5$ vs. $71.7$\% AP). The largest
improvement is observed for ankles: progressive prediction that conditions on the close-by parts in
the tree hierarchy reduces the distance between the conditioning signal and the location of the
predicted body part and simplifies the prediction task. Performing inference
(\tdbushort{}+\spatprop) improves the performance to $73.3$\% AP, due to more optimal assignment of
part detection candidates to corresponding persons. Graph simplification in \tdbushort{}
 allows to further reduce the inference time for graph partitioning ($0.08$ vs. $0.22$ for
$\busparse$).

\tabcolsep 1.5pt
\begin{table}[tbp]
 \scriptsize
  \centering
  \begin{tabular}{@{} l c ccc ccc cc@{}}
    \toprule
    Setting& Head   & Sho  & Elb & Wri & Hip & Knee & Ank & AP \\
    \midrule
    \tdshort & 91.6  & 84.7  & \textbf{72.9}  & \textbf{63.2}  & 72.3  & 64.7 & 52.8 & 71.7 \\ 
    \tdshort{} + \spatprop & 90.7  & 85.0  & 72.0  & 63.1  & 73.1  & 65.0 & 58.3 & 72.5 \\ 
    \tdbushort{} + \spatprop & \textbf{92.2}  & \textbf{86.1}  & 72.8  & 63.0  & \textbf{74.0}  & \textbf{66.2} & \textbf{58.4} & \textbf{73.3} \\

    \bottomrule
  \end{tabular}
 \vspace{0.75em}
\caption[]{Effects of various versions of $\detection$ model on pose estimation performance (AP) on MPII Multi-Person Val.}
\label{tab:mpii-single-frame-val:detection}
\end{table}

\tabcolsep 1.5pt
\begin{table}[tbp]
 \scriptsize
  \centering
  \begin{tabular}{@{} l c ccc ccc ccc@{}}
    \toprule
    Setting& Head   & Sho  & Elb & Wri & Hip & Knee & Ank  & AP & \timeinfer \\
    \midrule
    
    \bufull & \textbf{91.5}  & \textbf{87.8}  & 74.6  & 62.5  & 72.2  & 65.3 & 56.7 & 72.9 & 0.12\\ 
    \tdbushort + \spatprop & 88.8  & 87.0  & \textbf{75.9}  & \textbf{64.9}  & \textbf{74.2}  & \textbf{68.8} & \textbf{60.5} & \textbf{74.3}  & \textbf{0.005}\\ 

    \midrule
    \deepercut~\cite{insafutdinov16eccv} & 79.1  & 72.2  & 59.7 & 50.0 & 56.0  & 51.0 & 44.6 & 59.4 & 485 \\ 
    \deepercut~\cite{insafutdinov16arxiv} & 89.4  & 84.5  & 70.4  & 59.3  & 68.9  & 62.7 & 54.6 & 70.0 & 485\\
    Iqbal\&Gall~\cite{Iqbal_ECCVw2016} & 58.4  & 53.9  & 44.5  & 35.0  & 42.2  & 36.7 & 31.1 & 43.1 & 10\\
    \bottomrule
  \end{tabular}
 \vspace{0.75em}

\caption[]{Pose estimation results (AP) on MPII Multi-Person Test.}
  \label{tab:mpii-single-frame:sota}
  \vspace{-1.7em}
\end{table}

\begin{figure*}
  \centering
  \begin{tabular}{c c c c c}

  \begin{sideways}\bf \small~~~Single Frame \end{sideways}&
  	\includegraphics[height=0.129\linewidth]{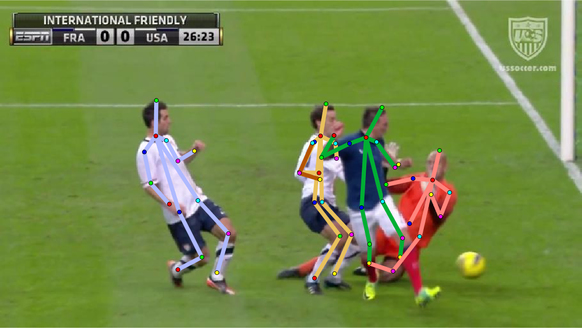}&
  	\includegraphics[height=0.129\linewidth]{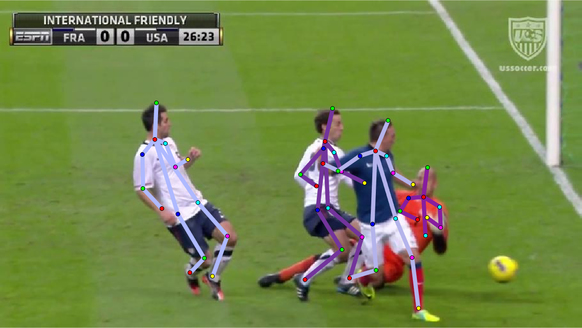}&
  	\includegraphics[height=0.129\linewidth]{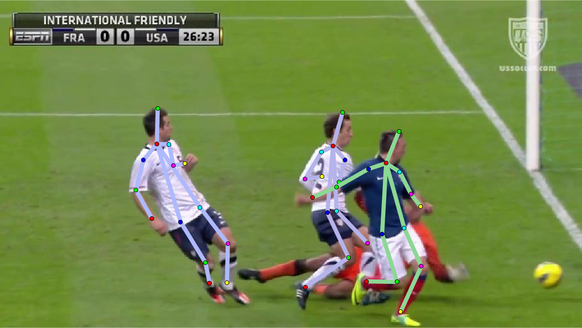}&
  	\includegraphics[height=0.129\linewidth]{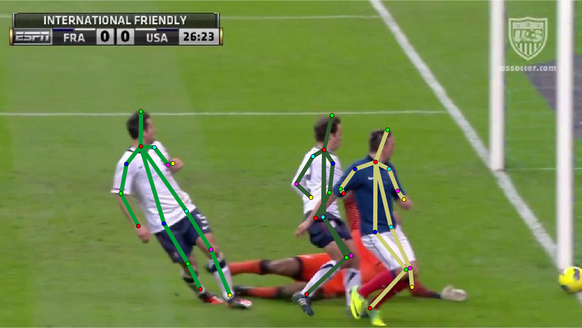}\\

  \begin{sideways}\bf \small~~~~~~Tracking \end{sideways}&
  	\includegraphics[height=0.129\linewidth]{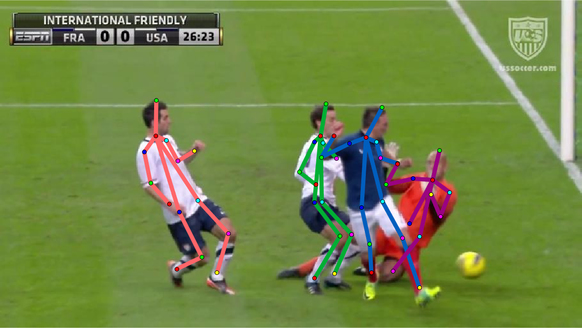}&
  	\includegraphics[height=0.129\linewidth]{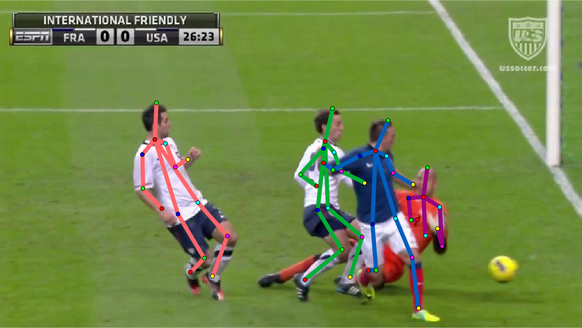}&
  	\includegraphics[height=0.129\linewidth]{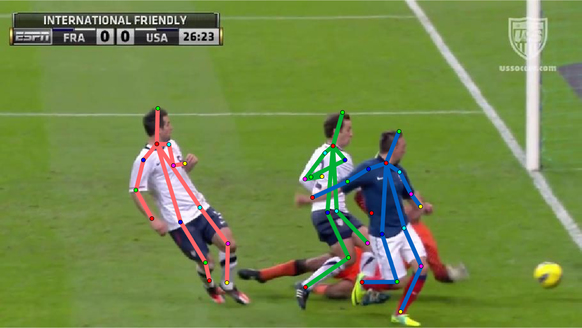}&
  	\includegraphics[height=0.129\linewidth]{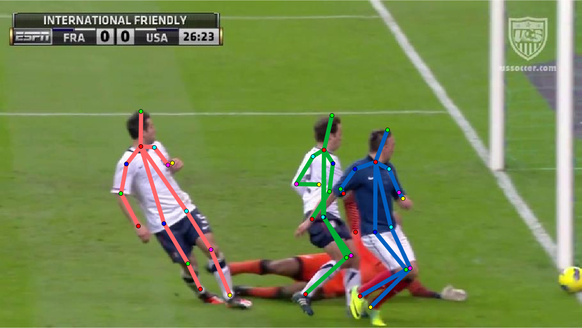}\\\vspace{0.1cm}
    ~& (a) & (b)  & (c) & (d) \\
  \begin{sideways}\bf \small~~~Single Frame \end{sideways}&
  	\includegraphics[height=0.129\linewidth]{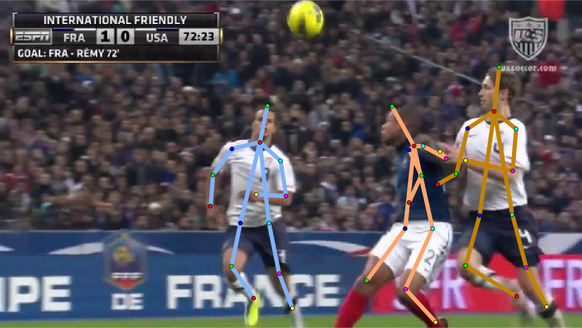}&
  	\includegraphics[height=0.129\linewidth]{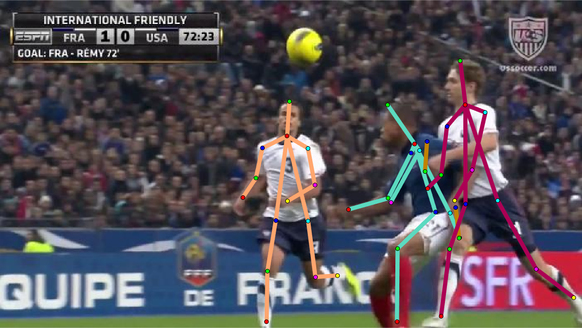}&
  	\includegraphics[height=0.129\linewidth]{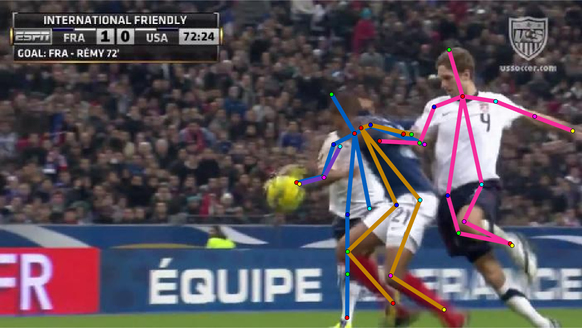}&
  	\includegraphics[height=0.129\linewidth]{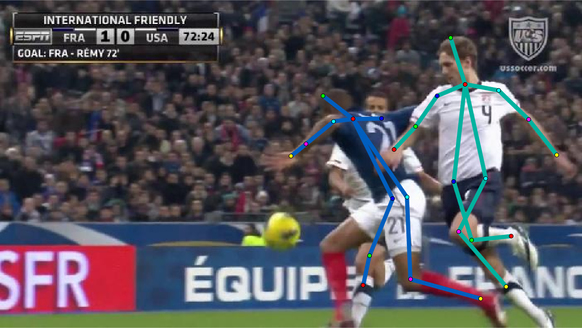}\\

  \begin{sideways}\bf \small~~~~~~Tracking \end{sideways}&
  	\includegraphics[height=0.129\linewidth]{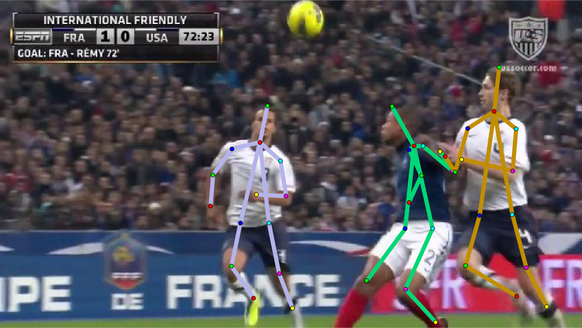}&
  	\includegraphics[height=0.129\linewidth]{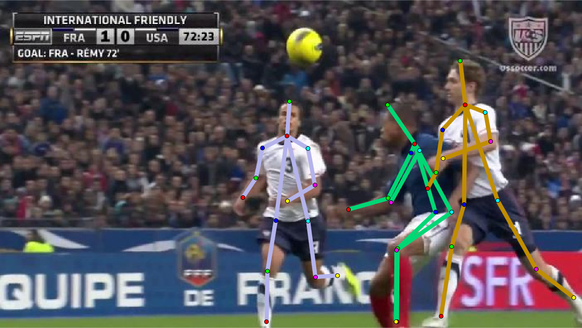}&
  	\includegraphics[height=0.129\linewidth]{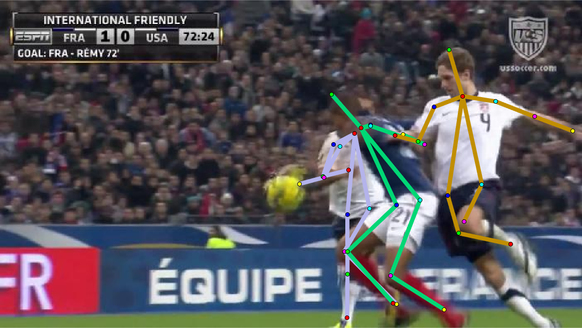}&
  	\includegraphics[height=0.129\linewidth]{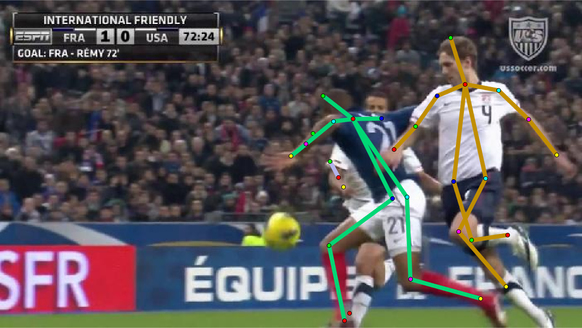}\\\vspace{0.1cm}
    ~& (e) & (f)  & (g) & (h) \\


  \end{tabular}
   \caption{Qualitative comparison of results using single frame based
     model (\busparse{}) vs. articulated tracking (\busparsevid{}). See \url{http://youtube.com/watch?v=eYtn13fzGGo} for the supplemental material showcasing our results.}
  \label{fig:qualitative_video}
\end{figure*}


\myparagraph{Comparison to the State of the Art.} We compare the
proposed single-frame approaches to the state of the art on MPII
Multi-Person Test and WAF~\cite{eichner10eccv} datasets. Comparison on
MPII is shown in Tab.~\ref{tab:mpii-single-frame:sota}. Both $\bufull$
and $\tdbushort$ improve over the best published result of
\deepercut{}~\cite{insafutdinov16arxiv}, achieving $72.9$ and $74.3$\%
AP respectively vs. $70.0$\% AP by \deepercut{}. For the $\tdbushort$
the improvements on articulated parts (elbows, wrists, ankles, knees)
are particularly pronounced. We argue that this is due to using the network
that is directly trained to disambiguate body parts of different
people, instead of using explicit geometric pairwise terms that only
serve as a proxy to person's identity. Overall, the performance of our
best $\tdbushort$ method is noticeably higher ($74.3$ vs. $70.0$\%
AP). Remarkably, its run-time \timeinfer~of graph partitioning stage is $5$ orders of magnitude
faster compared to $\deepercut$. This speed-up is due to two
factors. First, $\tdbushort$ relies on a faster solver
\cite{levinkov16arxiv} that tackles the graph-partitioning problem via
local search, in contrast to the exact solver used in
\cite{insafutdinov16eccv}. Second, in the case of $\tdbushort$ model
the graph is sparse and a large portion of the computation is
performed by the feed-forward CNN introduced in
\Sec.~\ref{sec:person-cond}. On WAF~\cite{eichner10eccv} dataset
\tdbushort~substantially improves over the best published result
($87.7$ vs. $82.0$\% AP by~\cite{insafutdinov16arxiv}). We refer to
supplemental material for details.



\subsection{Multi-frame models}

\myparagraph{Comparison of video-based models.} Performance of the
proposed video-based models is compared in
Tab.~\ref{tab:mpii-multi-video:overall}. Video-based models outperform
single-frame models in each case. $\bufullvid$ slightly outperforms
$\bufull$, where improvements are noticeable for ankle, knee and
head. $\busparsevid$ noticeably improves over $\busparse$ ($73.1$
vs. $71.6$\% AP). We observe significant improvements on the most
difficult parts such as ankles ($+3.9$\% AP) and wrists ($+2.6$\%
AP). Interestingly, $\busparsevid$ outperforms $\bufull+\temporal$:
longer-range connections such as, \eg, head to ankle, may introduce
additional confusion when information is propagated over
time. Finally, $\tdbuvid$ improves over $\tdbushort$ ($+0.7$\%
AP). Similarly to $\busparsevid$, improvement is most prominent on
ankles ($+1.8$\% AP) and wrists ($+0.9$\% AP). Note that even the
single-frame \tdbushort{} outperforms the best temporal \bushort{}
model. 
We show examples of articulated tracking on \videodata~in Fig.~\ref{fig:qualitative_video}. Temporal reasoning helps in cases
when image information is ambiguous due to close proximity of multiple
people. For example the video-based approach succeeds in correctly
localizing legs of the person in Fig.~\ref{fig:qualitative_video} (d)
and (h). 

\myparagraph{Temporal features.} We perform an ablative experiment on the \videodata~ dataset to
evaluate the individual contribution of the temporal features introduced in
Sec.~\ref{subsection:temporal}. The Euclidean distance alone achieves $72.1$ AP, adding DeepMatching
features improves the resuls to $72.5$ AP, whereas the combination of all features achieves the best result
of $73.1$ AP (details in supplemental material).


\tabcolsep 1.5pt
\begin{table}[tbp]
 \scriptsize
  \centering
  \begin{tabular}{@{} l c ccc ccc ccc@{}}
    \toprule
    Setting& Head   & Sho  & Elb & Wri & Hip & Knee & Ank & AP \\
    \midrule
    \bufull & 84.0  & 83.8  & 73.0  & 61.3  & 74.3  & 67.5 & 58.8 & 71.8 \\
    \quad + \temporal & 84.9  & 83.7  & 72.6  & 61.6  & 74.3  & 68.3 & 59.8 & 72.2  \\

    \midrule
    \busparse & 84.5  & 84.0  & 71.8  & 59.5  & \textbf{74.4}  & 68.1 & 59.2 & 71.6 \\
    \quad + \temporal & \textbf{85.6}  & 84.5  & 73.4  & 62.1  & 73.9  & 68.9 & 63.1 & 73.1 \\ 
   
    \midrule
    \tdbushort + \spatprop & 82.2  & 85.0  & 75.7  & 64.6  & 74.0  & 69.8 & 62.9 & 73.5 \\ 
    \quad + \temporal      & 82.6  & \textbf{85.1}  & \textbf{76.3}  & \textbf{65.5}  & 74.1  & \textbf{70.7} & \textbf{64.7} & \textbf{74.2} \\ 
    \bottomrule
  \end{tabular}
 \vspace{0.75em}
\caption[]{Pose estimation results (AP) on \videodata.} 
  \label{tab:mpii-multi-video:overall}
\end{table}


\myparagraph{Tracking evaluation.} In Tab.~\ref{tab:mpii-multi-video:mota} we present results of the
evaluation of multi-person articulated body tracking. We treat each body joint of each person as a
tracking target and measure tracking performance using a standard multiple object tracking accuracy
(MOTA) metric \cite{Bernardin:2008:CLE} that incorporates identity switches, false positives and
false negatives\footnote{Note that MOTA metric does not take the confidence scores of detection or
  track hypotheses into account. To compensate for that in the experiment in
  Tab.~\ref{tab:mpii-multi-video:mota} we remove all body part detections with a score 
  $\leq 0.65$ for \busparse~and $\leq 0.7$ for \tdbushort~prior to evaluation.}. We experimentally compare to a baseline model that first tracks people
across frames and then performs per-frame pose estimation.
%
%
To track a person we use a reduced version of our algorithm that operates on the two head
joints only. This allows to achieve near perfect person tracking results in most cases. Our tracker still fails
when the person head is occluded for multiple frames as it does not incorporate long-range
connectivity between target hypothesis. We leave handling of long-term occlusions for the future
work.
%
%
For full-body tracking we use the same inital head tracks and add them to the set of body part
proposals, while also adding must-link and must-cut constraints for the temporal edges corresponding
to the head parts detections.
%
%
The rest of the graph remains unchanged so that at inference time the body parts can be freely
assigned to different person tracks. For the \busparse{}~the full body tracking improves performance
by $+5.9$ and $+5.8$ MOTA on wrists and ankles, and by $+5.0$ and $+2.4$ MOTA on elbows and knees
respectively. \tdbushort~benefits from adding temporal connections between body parts as well, but
to a lesser extent than \busparse{}. The most significant improvement is for ankles ($+1.4$
MOTA). \busparse{}~also achieves the best overall score of $58.5$ compared to $55.9$ by \tdbushort.
 This is surprising since \tdbushort~outperformed \busparse~on the pose
estimation task (see Tab.~\ref{tab:mpii-single-frame-val:regression} and
\ref{tab:mpii-single-frame:sota}). We hypothesize that limited improvement of \tdbushort~could be
due to balancing issues between the temporal and spatial pairwise terms that are estimated
independently of each other.
%
%
%
\tabcolsep 1.5pt
\begin{table}[tbp]
 \scriptsize
  \centering
  \begin{tabular}{@{} l c ccc ccc ccc@{}}
    \toprule
    Setting & Head   & Sho  & Elb & Wri & Hip & Knee & Ank & Average \\
    \midrule

  Head track + \busparse  & 70.5  & 71.7  & 53.0  & 41.7  & 57.0 & 52.4 & 41.9 & 55.5 \\
  \quad \quad + \temporal & \textbf{70.6}  & \textbf{72.7}  & \textbf{58.0}  & \textbf{47.6}  & \textbf{57.6} & \textbf{54.8} & \textbf{47.7} & \textbf{58.5} \\

    
    \midrule

   Head track + \tdbushort & 64.8  & 69.4  & 55.4  & 43.4  & 56.4 & 52.2 & 44.8 & 55.2 \\
   \quad \quad + \temporal & 65.0  & 69.9  & 56.3  & 44.2  & 56.7 & 53.2 & 46.1 & 55.9 \\

    
    \bottomrule
  \end{tabular}
 \vspace{0.75em}
\caption[]{Tracking results (MOTA) on the \videodata.} 
  \label{tab:mpii-multi-video:mota}
  \vspace{-1.5em}
\end{table}



 
\section{Conclusion}
In this paper we introduced an efficient and effective approach to
articulated body tracking in monocular video. Our
approach defines a model that jointly groups body part proposals
within each video frame and across time. Grouping is formulated as a
graph partitioning problem that lends itself to efficient inference
with recent local search techniques. Our approach improves over
state-of-the-art while being substantially faster compared to other
related work.\\\vspace{0.1em}

\myparagraph{Acknowledgements.} This work has been supported by the Max Planck Center for Visual
Computing and Communication. The authors thank Varvara Obolonchykova
and Bahar Tarakameh for their help in creating the video dataset.

\begin{appendices}
\section{Additional Results on the MPII Multi-Person Dataset}

\begin{figure*}
	\centering
	\begin{tabular}{c c c c c}
	    \begin{sideways} \qquad \qquad  BU \end{sideways}&		
		\includegraphics[width=0.2361\linewidth]{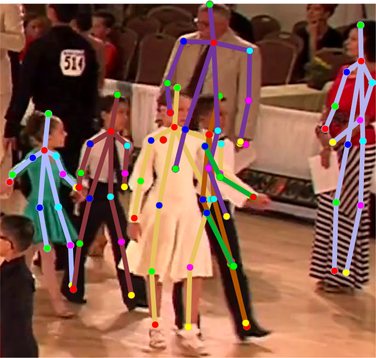}&
		\includegraphics[width=0.2215\linewidth]{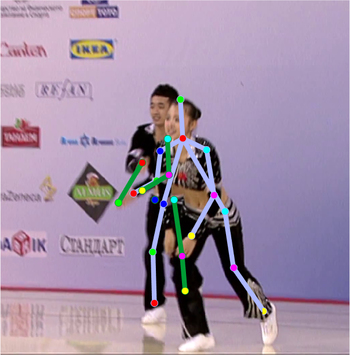}&
		\includegraphics[width=0.2548\linewidth]{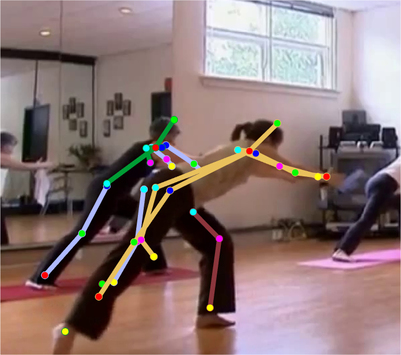}&
		\includegraphics[width=0.2153\linewidth]{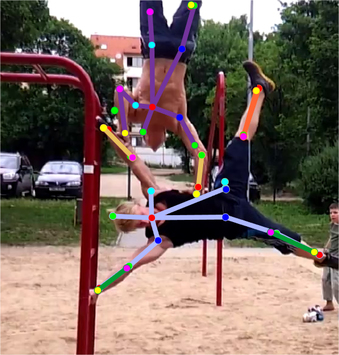}\\

	    \begin{sideways} \qquad \qquad TD/BU \end{sideways}&		
		\includegraphics[width=0.2361\linewidth]{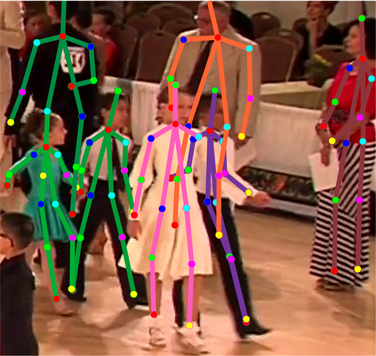}&
		\includegraphics[width=0.2215\linewidth]{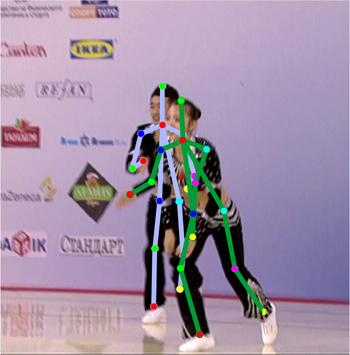}&
		\includegraphics[width=0.2548\linewidth]{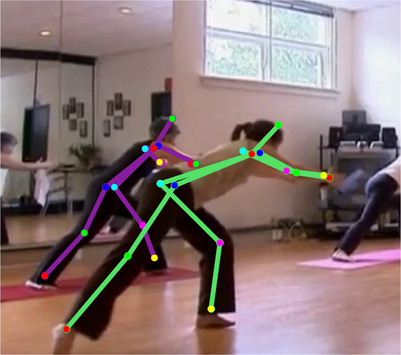}&
		\includegraphics[width=0.2153\linewidth]{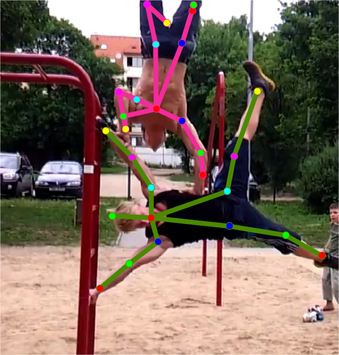}\\
        & 1 & 2 & 3 & 4 \\
        &  &  &  & \qquad \\
  	\end{tabular}
	
	\begin{tabular}{c c c c}
	    \begin{sideways} \qquad \qquad \quad  BU \end{sideways}&		
	    \includegraphics[width=0.35\linewidth]{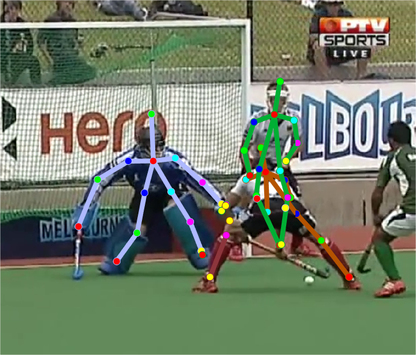}&
		\includegraphics[width=0.28\linewidth]{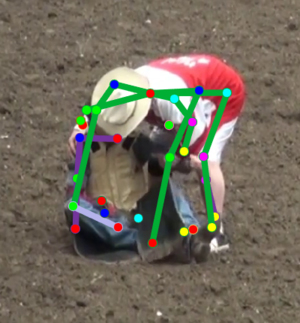}&
		\includegraphics[width=0.31\linewidth]{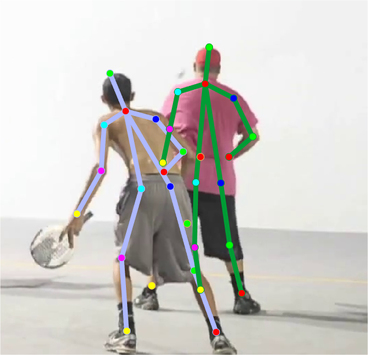}\\
		
	    \begin{sideways} \qquad \qquad \quad TD/BU \end{sideways}&		
	    \includegraphics[width=0.35\linewidth]{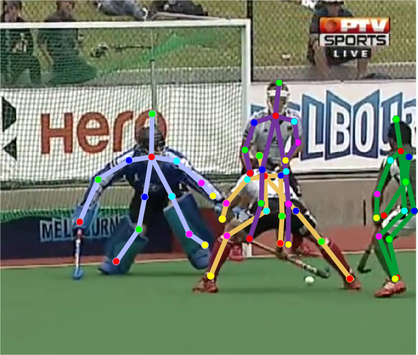}&
		\includegraphics[width=0.28\linewidth]{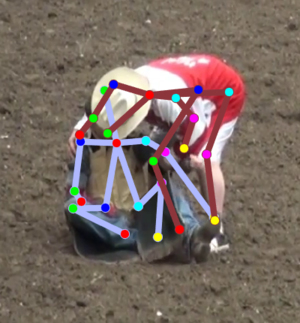}&
		\includegraphics[width=0.31\linewidth]{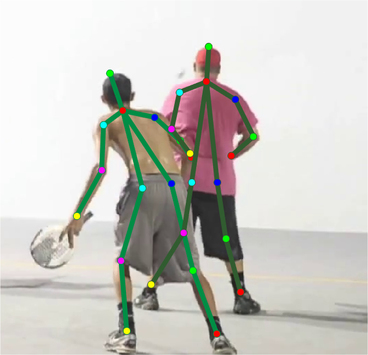}\\
        & 5 & 6 & 7 \\
        &  &  & \qquad \\
	\end{tabular}
    \caption{Qualitative comparison of single-frame based \tdbushort{}
      and \bufull{} on MPII Multi-Person dataset.
    }
	\label{fig:qualitative_mpii_supp}

\end{figure*}

\begin{figure*}
	\centering
	
	\begin{tabular}{c c c c}
		
		\begin{sideways} \qquad \qquad \quad  BU \end{sideways}&
        \includegraphics[height=0.20\linewidth]{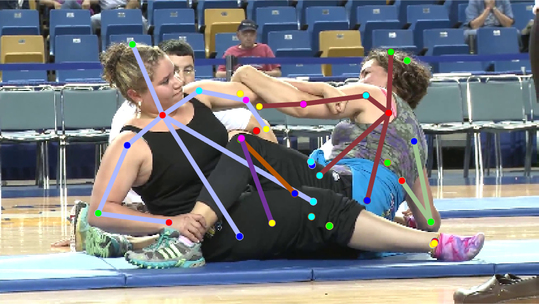}&
		\includegraphics[height=0.20\linewidth]{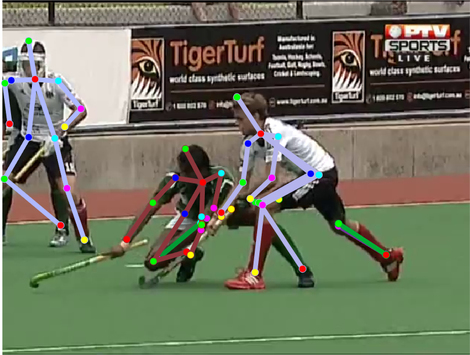}&
		\includegraphics[height=0.20\linewidth]{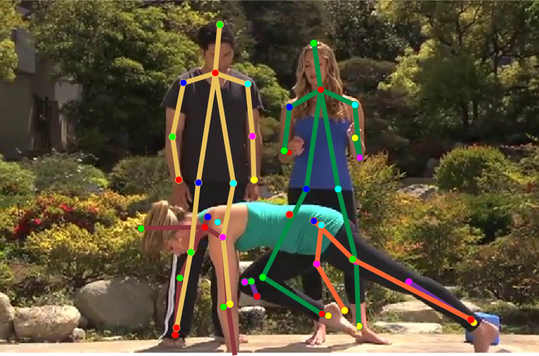}\\

		\begin{sideways} \qquad \qquad \quad TD/BU \end{sideways}&
        \includegraphics[height=0.20\linewidth]{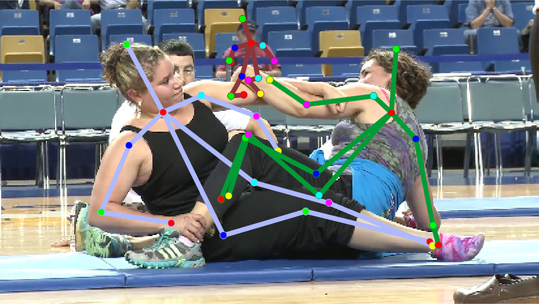}&
        \includegraphics[height=0.20\linewidth]{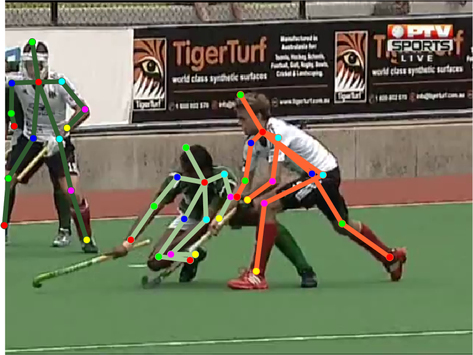}&
		\includegraphics[height=0.20\linewidth]{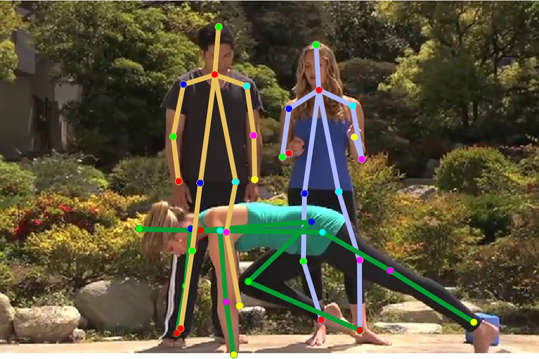}\\

        & 8 & 9 & 10 \\
        &  &  & \qquad \\		
	\end{tabular}
	
	\begin{tabular}{c c c c}
		
		\begin{sideways} \qquad \qquad \quad  BU \end{sideways}&
		\includegraphics[height=0.22\linewidth]{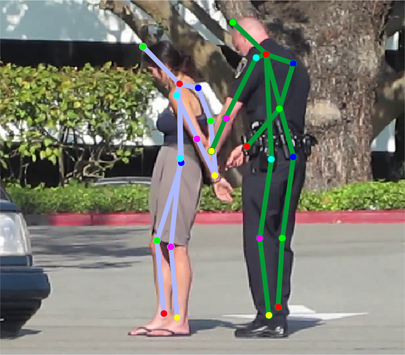}&
		\includegraphics[height=0.22\linewidth]{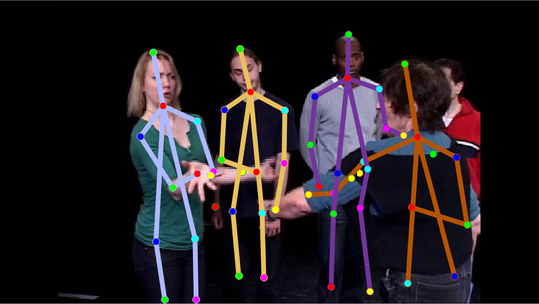}&
		\includegraphics[height=0.22\linewidth]{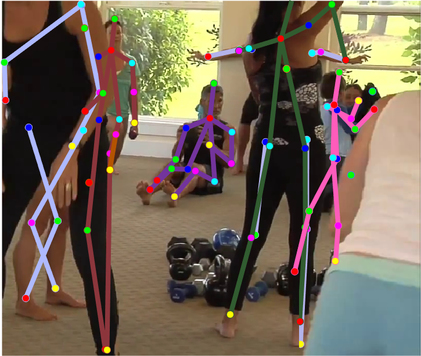}\\
						
		\begin{sideways} \qquad \qquad \quad TD/BU \end{sideways}&
		\includegraphics[height=0.22\linewidth]{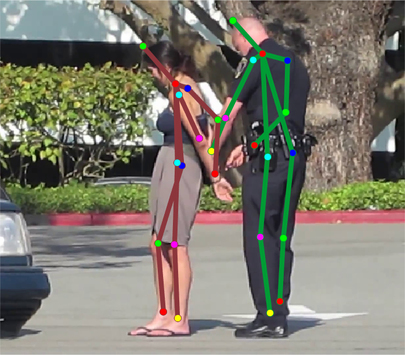}&
		\includegraphics[height=0.22\linewidth]{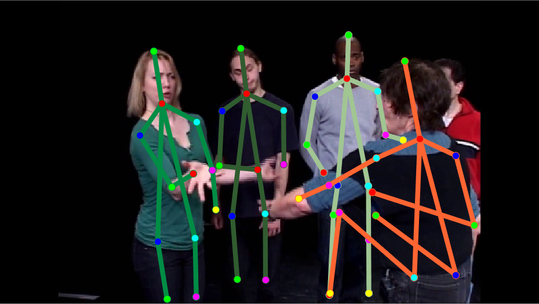}&
		\includegraphics[height=0.22\linewidth]{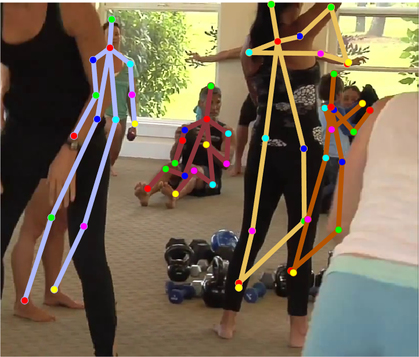}\\
        & 11 & 12 & 13 \\
        &  &  & \qquad \\	
	\end{tabular}
	
    \caption{Successfull (8-11) and failure (12-13) pose estimation
      results by single-frame based \tdbushort{} and comparison to
      \bufull{} on MPII Multi-Person dataset.}
    \label{fig:qualitative_mpii_supp_2}

\end{figure*}

We perform qualitative comparison of the proposed single-frame based
~\tdbushort{} and \bufull{} methods on challenging scenes containing
highly articulated and strongly overlapping individuals. Results are
shown in Fig.~\ref{fig:qualitative_mpii_supp} and
Figure~\ref{fig:qualitative_mpii_supp_2}.
The \bufull{} works well when persons are sufficiently separated (images
11 and 12). However, it fails on images where people significantly
overlap (images 1-3, 5-10) or exhibit high degree of articulation
(image 4). This is due to the fact that geometric image-conditioned
pairwise may get confused in the presence of multiple overlapping
individuals and thus mislead post-CNN bottom-up assembling of body
poses. In contrast, \tdbushort{} performs explicit modeling of person
identity via top-dop bottom-up reasoning while offloading the larger
share of the reasoning about body-part association onto feed-forward
convolutional architecture, and thus is able to resolve such
challenging cases. Interestingly, \tdbushort{} is able to correctly
predict lower limbs of people in the back through partial occlusion
(image 3, 5, 7, 10). \tdbushort{} model occasionally incorrectly
assembles body parts in kinematically implausible manner (image 12),
as it does not explicitly model geometric body part
relations. Finally, both models fail in presense of high variations in
scale (image 13). We envision that reasoning over multiple scales is
likely to improve the results.

\section{Results on the We Are Family dataset}

\tabcolsep 1.5pt
\begin{table}[tbp]
 \scriptsize
  \centering
  \begin{tabular}{@{} l c ccc c@{}}
    \toprule
    Method & Head   & Sho  & Elb & Wri & Total \\
    \midrule
    \tdbushort  & \textbf{97.5}  & \textbf{86.2}  & \textbf{82.1}  & \textbf{85.2} & \textbf{87.7} \\
    
    \midrule
    $\deepercut$ \cite{insafutdinov16eccv}  & 92.6  & 81.1  & 75.7  & 78.8 & 82.0 \\ 
    $\deepcut$~\cite{pishchulin16cvpr}& 76.6  & 80.8  & 73.7  & 73.6  & 76.2 \\
    Chen\&Yuille~\cite{Chen:2015:POC}  & 83.3  & 56.1 & 46.3  & 35.5 & 55.3 \\

    \bottomrule
  \end{tabular}
  \caption[]{Pose estimation results (AP) on WAF dataset.}
    \vspace{-1.0em}
  \label{tab:multicut:waf:ap}
\end{table}


\tabcolsep 1.5pt
\begin{table}[tbp]
 \scriptsize
  \centering
  \begin{tabular}{@{} l c ccc ccc cc@{}}
    \toprule
    Setting& Head  & Sho  & Elb & Wri & Hip & Knee & Ank & AP \\
    \midrule
    \busparse & 84.5  & 84.0  & 71.8  & 59.5  & 74.4  & 68.1 & 59.2 & 71.6 \\ 
    \quad + \detdistance & 84.8  & 84.3  & 72.9  & 61.8  & 74.1  & 67.4 & 59.1 & 72.1 \\ 
    \quad + \deepmatch & 85.5  & 83.9  & 73.0  & 62.0  & 74.0  & 68.0 & 59.5 & 72.3 \\ 
    \quad\quad + \detdistance & 85.1  & 83.6  & 72.2  & 61.5  & \textbf{74.4}  & 68.8 & 62.2 & 72.5 \\ 
    \quad\quad\quad + \siftdistance & \textbf{85.6}  & \textbf{84.5}  & \textbf{73.4}  & \textbf{62.1}  & 73.9  & \textbf{68.9} & \textbf{63.1} & \textbf{73.1} \\ 
    \bottomrule
  \end{tabular}
 \vspace{0.75em}
\caption[]{Effects of different temporal features on pose estimation performance (AP) ($\busparsevid$ model) on our \videodata.}
  \label{tab:mpii-multi-video:features}
  \vspace{-1.5em}
\end{table}

We compare our proposed \tdbushort{} model to the state-of-the-art methods on the
``We Are Family'' (WAF)~\cite{eichner10eccv} dataset and present results in
Table~\ref{tab:multicut:waf:ap}. We use evaluation protocol from \cite{insafutdinov16eccv}
and report the AP evaluation measure. \tdbushort{} model outperforms the best published results
\cite{insafutdinov16eccv} across all body parts ($87.7$\ vs $82.0$\% AP)
as well improves on articulated parts such as wrists ($+6.4$\% AP)
and elbows ($+6.4$\% AP). We attribute that to the ability of top-down model to better
learn part associations compared to explicit modeling geometric pairwise relations
as in \cite{insafutdinov16eccv}.

\section{Evaluation of temporal features.}

We evaluate the importance of combining temporal
features introduced in Sec. 3.4 of the paper on our Multi-Person Video dataset. To that end,
we consider $\busparsevid$ model and compare results to $\busparse$ in
Tab.~\ref{tab:mpii-multi-video:features}. Single-frame $\busparse$ achieves $71.6$\% AP. It
can be seen that using geometry based $\detdistance$ features slightly improves the results to
$72.1$\% AP, as it enables the propagation of information from neighboring frames. Using $\deepmatch$
features slightly improves the performance further as it helps to link the same body part of the
same person over time based on the body part appearance. It is especially helpful in the case of fast
motion where $\detdistance$ may fail. The combination of both geometry and appearance based features
further improves the performance to $72.5$\%, which shows their complementarity. Finally, adding
the $\siftdistance$ feature improves the results to $73.1$\%, since it copes better with the
sudden changes in background and body part orientations. Overall, using a combination of temporal
features in \busparsevid{} results in a $1.5$\% AP improvement over the single-frame
\busparse.
This demonstrates the advantages of the proposed approach to improve pose estimation performance
using temporal information.
\end{appendices}
{\small
\bibliographystyle{ieee}
\bibliography{biblio}
}

\end{document}